\documentclass[journal]{IEEEtran}
\usepackage{cite}

\ifCLASSINFOpdf
   \usepackage[pdftex]{graphicx}
 \graphicspath{{../pdf/}{../jpeg/}}
\DeclareGraphicsExtensions{.pdf,.jpeg,.png}
\else
\usepackage[dvips]{graphicx}
\graphicspath{{../eps/}}
 \DeclareGraphicsExtensions{.eps}
\fi
\usepackage{amsmath}
\usepackage{algorithmic}

\usepackage{array}

\ifCLASSOPTIONcaptionsoff
  \usepackage[nomarkers]{endfloat}
 \let\MYoriglatexcaption\caption
 \renewcommand{\caption}[2][\relax]{\MYoriglatexcaption[#2]{#2}}
\fi
\usepackage{url}

\usepackage{times}
\usepackage{latexsym}
\usepackage{amssymb}
\usepackage[normalem]{ulem}
\usepackage{color}
\usepackage{colortbl}
\usepackage{multirow}
\usepackage{caption}

\begin{document}

\newcolumntype{I}{!{\vrule width 0pt}}

%
\title{Robust Reading Comprehension with \\Linguistic Constraints via Posterior Regularization}
%
%
%

\author{Mantong~Zhou,
        Minlie~Huang,
        Xiaoyan~Zhu
\thanks{Mantong Zhou, Minlie Huang and Xiaoyan Zhu are with the Institute for Artificial Intelligence, 
State Key Lab of Intelligent Technology and Systems, 
Beijing National Research Center for Information Science and Technology, 
Department of Computer Science and Technology, Tsinghua University, Beijing 100084, China, e-mail: (zmt.keke@gmail.com, aihuang@tsinghua.edu.cn, zxy-dcs@tsinghua.edu.cn).}
\thanks{Corresponding author: Minlie Huang.}
}

%
%

\markboth{IEEE/ACM TRANSACTIONS ON AUDIO, SPEECH, AND LANGUAGE PROCESSING,~Vol.*, No.~*,* *}%
{Shell \MakeLowercase{\textit{et al.}}: Bare Demo of IEEEtran.cls for IEEE Journals}
%



\maketitle

\begin{abstract}
 In spite of great advancements of machine reading comprehension (RC), existing RC models are still vulnerable and not robust to different types of adversarial examples. 
 Neural models over-confidently predict wrong answers to semantic different adversarial examples, while over-sensitively predict wrong answers to semantic equivalent adversarial examples.
 Existing methods which improve the robustness of such neural models merely mitigate one of the two issues but ignore the other.
 In this paper, we address the {\it over-confidence} issue and the {\it over-sensitivity} issue existing in current RC models simultaneously with the help of external linguistic knowledge. 
 We first incorporate external knowledge to impose different linguistic constraints (entity constraint, lexical constraint, and predicate constraint), and then regularize RC models through posterior regularization.
 Linguistic constraints induce more reasonable predictions 
 for both semantic different and semantic equivalent adversarial examples, and posterior regularization provides an effective mechanism to incorporate these constraints. 
 Our method can be applied to any existing neural RC models including state-of-the-art BERT models.
 Extensive experiments show that our method remarkably improves the robustness of base RC models, and is better to cope with these two issues simultaneously.
\end{abstract}


 \ifCLASSOPTIONpeerreview
 \begin{center} \bfseries EDICS Category: 3-BBND \end{center}
 \fi
%
\IEEEpeerreviewmaketitle

\section{Introduction}
%
%
%
%

\label{sec:intro}

\begin{figure*}[ht]
    \centering
    \includegraphics[width=18cm]{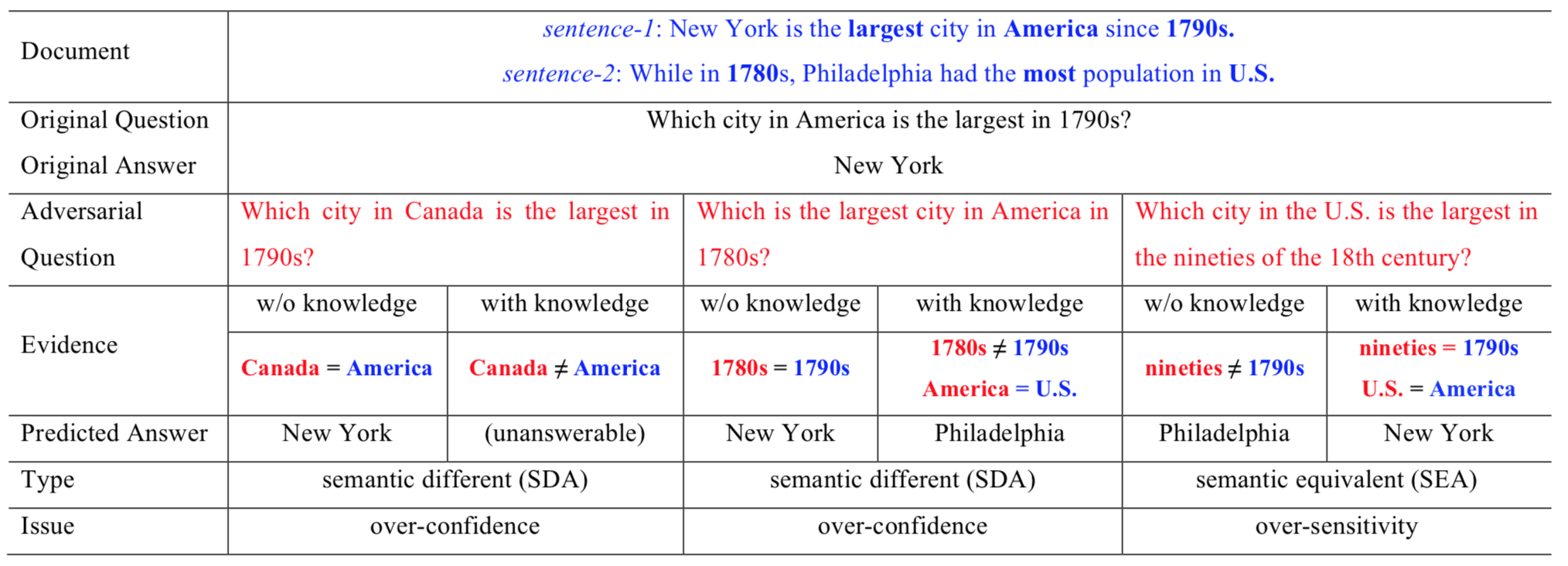}
    \caption{Different types of adversarial examples in reading comprehension.}
    \label{fig:example}
\end{figure*}

\IEEEPARstart{R}{eading} Comprehension (RC) has been much advanced by recently proposed datasets~\cite{CBT,squad,NarrativeQA} and models \cite{hu2018reinforced,sun2018u-net:}. 
However, RC models are still vulnerable and faced with two typical issues. 
One is \textbf{the over-confidence issue}: {\it when a model is fed with \textbf{S}emantic \textbf{D}ifferent \textbf{A}dversarial (\textbf{SDA}) examples}~\cite{squadadv}, {\it the model wrongly predicts the same answer}. If small perturbations are applied to the question/passage, for instance replacing ``America" with ``Canada" in the example of Fig.~\ref{fig:example}, the model still predicts the same answer even though the question is unanswerable.
The other is \textbf{the over-sensitivity issue}: {\it a model is not robust when fed with \textbf{S}emantic \textbf{E}quivalent \textbf{A}dversarial (\textbf{SEA}) examples}~\cite{sear}. If we make perturbations yet keep the semantics unchanged, for instance replacing ``1790s" with ``nineties of the 18th century", the model may be distracted and possibly predict wrong answers.

The over-confidence issue can be attributed to overfitting~\cite{calibration,pathologies}: the model, for instance, may use a trick to output low-entropy distributions over classes to minimize negative log-likelihood loss. 
The over-sensitivity issue can be attributed to non-local generalization of neural networks caused by massive nonlinear operations~\cite{nonlinear} or high-dimensional linear operations~\cite{advexp}. 
As for RC, the representation dilemma additionally leads to the over-confidence and over-sensitivity issues: if words are replaced with analogous but semantic-different counterparts (``America" $\rightarrow$ ``Canada", where the entities have very similar embeddings), the representation of the sentence may change slightly, so it is not surprising that a RC model outputs the same answer. 
Conversely, if words are replaced with synonymous phrases (``1790s" $\rightarrow$ ``nineties of the 18th century"), it is hard for the model to regard them as equivalent through much different representations. 
Recent solutions, such as entropy regularization~\cite{entropyreg} and adversarial training~\cite{advexp}, merely mitigate one issue but ignore the other. However, a robust RC model should be able to cope with both SDA and SEA examples simultaneously.

In this work, we aim to improve the robustness of RC models with linguistic constraints via posterior regularization (PR, \cite{pr}). 
Specifically, once we obtain predictions from a base RC model, we can extract linguistic feature pairs (synonyms, antonyms, entity pairs, etc.) with external knowledge resources, and derive linguistic constraints with the extracted features.
Then, we adjust the output distribution according to these linguistic constraints.
The training objective is reformulated as a constrained optimization problem in the posterior regularization framework, which can be solved by an expectation maximization (EM) algorithm.

The aforementioned issues are addressed as follows:
\textbf{First}, posterior regularization alleviates overfitting
by restricting the parameter space. Regularization term functions as penalty for vanilla negative log-likelihood loss.
\textbf{Second}, applying constraints to the output distribution is more straightforward than to the input or intermediate layers, which makes predictions less affected by massive non-linear operations.
\textbf{Third}, the constraints to regularize RC models are designed with paired linguistic knowledge (synonyms, antonyms,  entity pairs, etc.), which has two benefits. On one hand, linguistic constraints are designed to capture two types of adversarial examples simultaneously. On the other hand, instead of operating in the embedding space, symbolic changes in entity or lexicon can be more easily and explicitly captured by the constraints, and thus address the representation dilemma.

Our main contribution is to improve the robustness of reading comprehension models by \textbf{considering the over-confidence issue and the over-sensitivity issue simultaneously}.
We incorporate external linguistic knowledge to impose different constraints on the models via posterior regularization. 
Using this method, semantic different and semantic equivalent adversarial examples can be handled effectively. 
Moreover, our method can be applied to many RC models, and depends less on the quality of adversarial examples compared with adversarial training.

\section{Related Works}
\noindent \textbf{Adversarial Reading Comprehension Tasks}\\
Many studies~\cite{kaushik2018much,mudrakarta2018did} start to retrospect the benchmark datasets and tasks of reading comprehension (RC) critically. 
Existing RC models which perform well on SQuAD1.1~\cite{squad} are not robust to adversarial sentences. 
For instance, adversarial examples in SQuAD-ADDSENT~\cite{squadadv}, collected with semantic-altering noise using AddSent algorithm, fooled most of the successful models trained on SQuAD1.1. 
AddSentDiverse~\cite{wang2018robust} modified AddSent by generating more diverse adversarial examples to prevent RC models from learning superficial clues. 
Gao et al.~\cite{distractors} generated longer and semantic-richer distractors which are closer to those in real RC examinations.
Rajpurkar et al.~\cite{squad2.0} developed SQuAD2.0 that combines SQuAD1.1 with new unanswerable questions, to test the ability of distinguishing unanswerable questions in RC. 
SQuAD-ADDSENT and SQuAD2.0 examined the overconfidence issue of RC models using adversarial examples which differ in semantics. 

 Ribeiro et al.~\cite{sear} designed adversarial examples and rules, which can preserve original semantics but cause models to make wrong predictions. 
 Iyyer et al.~\cite{adversarialparaphrase} proposed syntactically controlled paraphrase networks to generate semantic equivalent adversarial examples. 

However, to our best knowledge, there exists no dataset which combines both SDA and SEA examples.

\noindent \textbf{Robust Reading Comprehension Models}\\
Current RC models are vulnerable since remarkable performance drops can be observed on adversarial examples compared to that on original examples. 
Entropy regularization~\cite{pathologies} is proposed to alleviate overfitting by maximizing the entropy of prediction distributions of similar inputs. 
Label smoothing~\cite{labelsmoothing} is equivalent to adding the KL divergence between the uniform distribution and the predicted answer distribution thus alleviates overfitting~\cite{entropyreg}.
But these methods only target at making models more sensitive to input permutations, yet ignoring the semantic-equivalent adversarial examples. 

The standard method to defend against adversarial attacks is adversarial training~\cite{mixdata,advexp}.
Szegedy et al.~\cite{mixdata} discovered that several machine learning models are vulnerable to adversarial examples and found that by training on a mixture of adversarial and clean examples, a neural network can be regularized to some degree.
Goodfellow et al.~\cite{advexp} introduced a family of fast methods for generating adversarial examples and demonstrated that adversarial training can result in regularization in theory.
Wang et al.~\cite{wang2018robust} introduced adversarial training to RC models and improved robustness using more diverse adversarial examples. 
A3Net~\cite{a3net} blended adversarial training into each layer of the network by adding numerical perturbations to original variables.
However, adversarial training relies on high-quality training adversarial examples~\cite{wang2018robust}. Such models cannot recognize unseen adversarial patterns 
without sufficient training data. 
In comparison, our model effectively
identifies adversarial patterns with the help of external knowledge.

Min et al.~\cite{sentenceselector} proposed a selector to pick out oracle sentences from adversarial ones, but cannot deal with unanswerable questions.
No-answer scoring~\cite{levy2017zero-shot,docqa} and answer verification modules~\cite{hu2019read,sun2018u-net:} were used in some models to determine whether a question is unanswerable. However, these modules depend on the specific ``no-answer-classification" setting of SQuAD2.0. 
By contrast, our framework is more general.

\begin{figure*}[!htp]
    \centering
    \includegraphics[width=16cm]{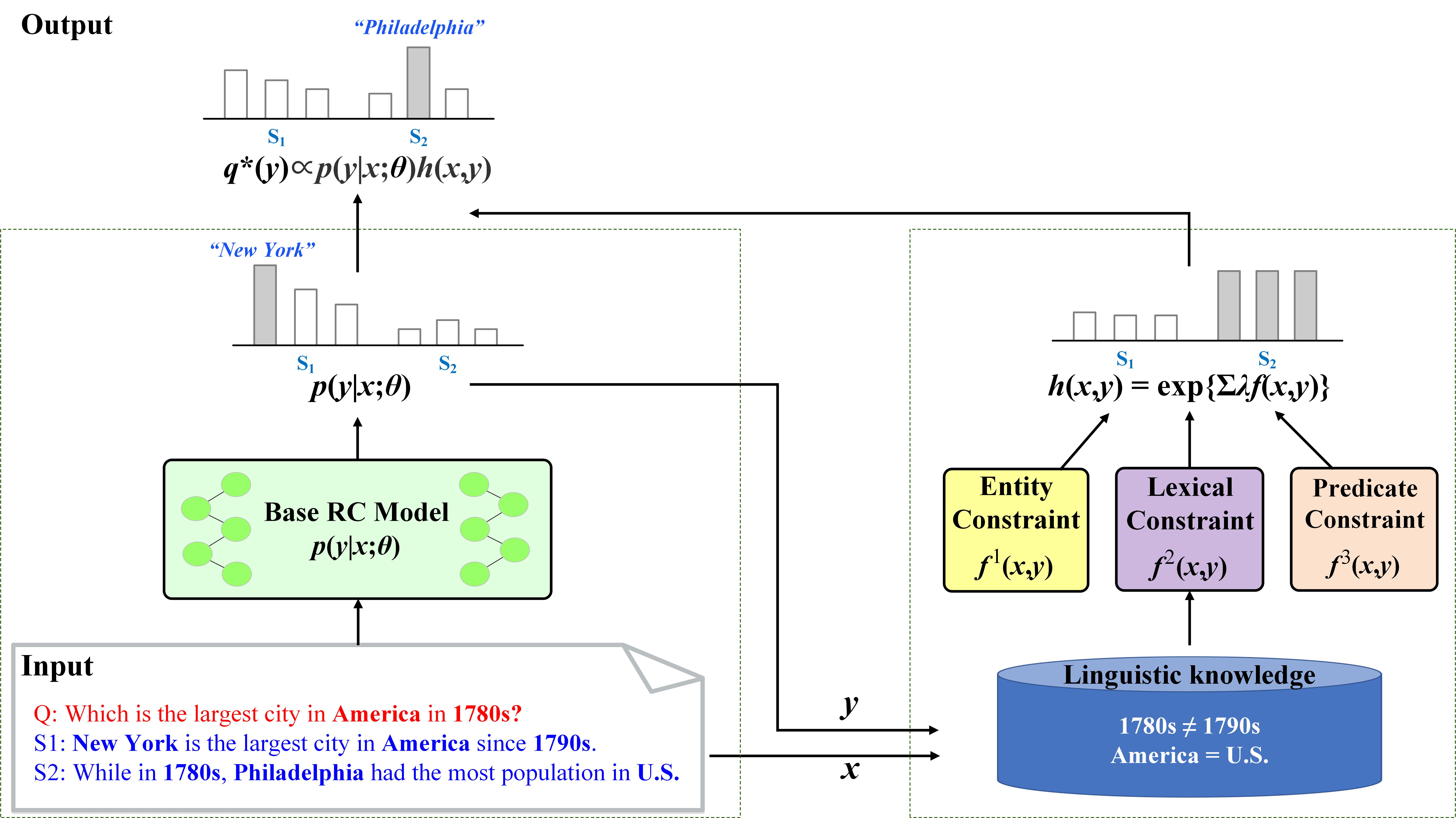}
    \caption{Posterior regularization framework for robust reading comprehension. Base RC model obtains the preliminary answer distribution $p(y|x;\theta)$, and the values of constraint functions $f^l(x,y)$ are computed via linguistic knowledge for each candidate answer. A larger $h(x,y)$ is derived for candidates which satisfy the constraints but a smaller $h(x,y)$ for those which violate the constraints. $h(x,y)$ is then used to regularize $p(y|x;\theta)$ to obtain the final answer distribution $q^*(y|x)$.}
    \label{fig:model}
\end{figure*}

\noindent \textbf{Posterior Regularization}\\
Posterior Regularization (PR)~\cite{pr} is a structured learning framework which enables flexible injection of various constraints with structured knowledge, and already applied to many NLP tasks such as Machine Translation~\cite{ganchev2013cross-lingual,Zhang2018Priormt,smtnmt} and Sentiment Classification~\cite{Yang2014Context,zhao2016semi}.
Hu et al.~\cite{logicnn} developed a knowledge distillation framework to incorporate PR into neural networks. A student neural network is trained to imitate a teacher network which is constructed by imposing posterior constraints. 
However, these models have limited generalization since constraints are fixed and manually designed.
Mei et al.~\cite{pmlr-v32-mei14} attempted to learn the constraint weights with additional supervisions in a Bayesian model with posterior regularization.
Hu et al.~\cite{learnedlogicnn} proposed mutual distillation to further enable the former distillation framework to learn constraints by parameterizing constraints.
These works inspired us to incorporate linguistic constraints to improve the robustness of RC models.

\section{Methodology}
\subsection{Reading Comprehension with Posterior Regularization}
The RC problem can be formulated as follows: given a set of triples $(Q,P,A)$, where $Q=(q_1,q_2,...,q_m)$ is the question with $m$ words, $P = (p_1,...,p_n)$ is the passage with $n$ words, and $A=(p_s,...,p_e)$ is the answer span extracted from the passage where $s$/$e$ indicates the start/end word position. The task is to build a model with parameters $\theta$ to estimate the conditional probability 
$p(A|Q,P;\theta)=p(s|Q,P;\theta)p(e|s,Q,P;\theta)$.

In general, we can optimize $\theta$ by maximizing the log-likelihood of the ground truth answer as follows:
\begin{align}
    &\max \mathcal{L}(\theta) =\sum_{i}\log p(A_i|Q_i,P_i;\theta) \notag \\
    &= \sum_{i} \log p(s_i|Q_i,P_i;\theta)
    +\log p(e_i|s_i,Q_i,P_i;\theta)
\end{align}

Our central idea is to build a robust model with linguistic constraints. Following the posterior regularization (PR) framework, we apply a set of constraints to the posterior distribution over the answers. We can define the constraints in the form of $E_q[\phi(x,y)]\leq b$ where $x$ and $y$ are an input question\footnote{We input a question-passage pair as $x$ to the RC model $p(y|x;\theta)$, but merely use the question in constraints $\phi(x,y)$.} 
and the answer respectively. $\phi(x,y)$ is a constraint function whose value is expected to be less than $b$ according to some particular linguistic rules, and $q$ is any distribution satisfying the constraints. 
The PR objective with slack penalty variables is as follows:
\begin{align}
    \max \mathcal{J}(\theta,q) &=\mathcal{L}(\theta)- \min_{q}\{ KL(q(y|x)|| p(y|x;\theta)) \notag \\  
    &+ C\sum\xi\}\\ 
    s.t. \quad &\lambda^{l}E_q[\phi^l(x,y)] - b^l \leq \xi \quad l=1,...,L \notag
\end{align}

Let $f^l(x,y)=b^l-\phi^l(x,y)$ denotes constraint functions whose values are expected to be larger than $-\xi$, i.e. $f^l(x,y)>0$, when $(x,y)$ satisfies the constraints.
The solution to the second term of $\mathcal{J}(\theta,q)$ is given by:
\begin{align}
    q^*(y|x)= \frac{p(y|x;\theta)\exp\{C\sum_l\lambda^lf^l(x,y)\}}{Z}
    \label{eq:q}
\end{align}
where $Z$ is the normalization factor. $q^*(y|x)$ is the desired distribution which is close to the distribution learned from data $p(y|x;\theta)$ and meanwhile is regularized by constraints.

In theory, this PR framework can be applied to any probabilistic model $p(y|x;\theta)$ which is called the base model hereafter.

\subsection{Linguistic Constraints}
We design three constraints in this section. 
\textbf{All constraints are designed to account for two types (semantic-different and semantic-equivalent) of adversarial examples in this framework}.
Since it is inadequate to pre-define all adversarial situations by deterministic functions,\textbf{ we introduce learnable parameters $\omega$ to parameterize the constraint functions as $f^l(x,y;\omega_l)$}.


\begin{figure}
    \centering
    \includegraphics[width=0.95\linewidth]{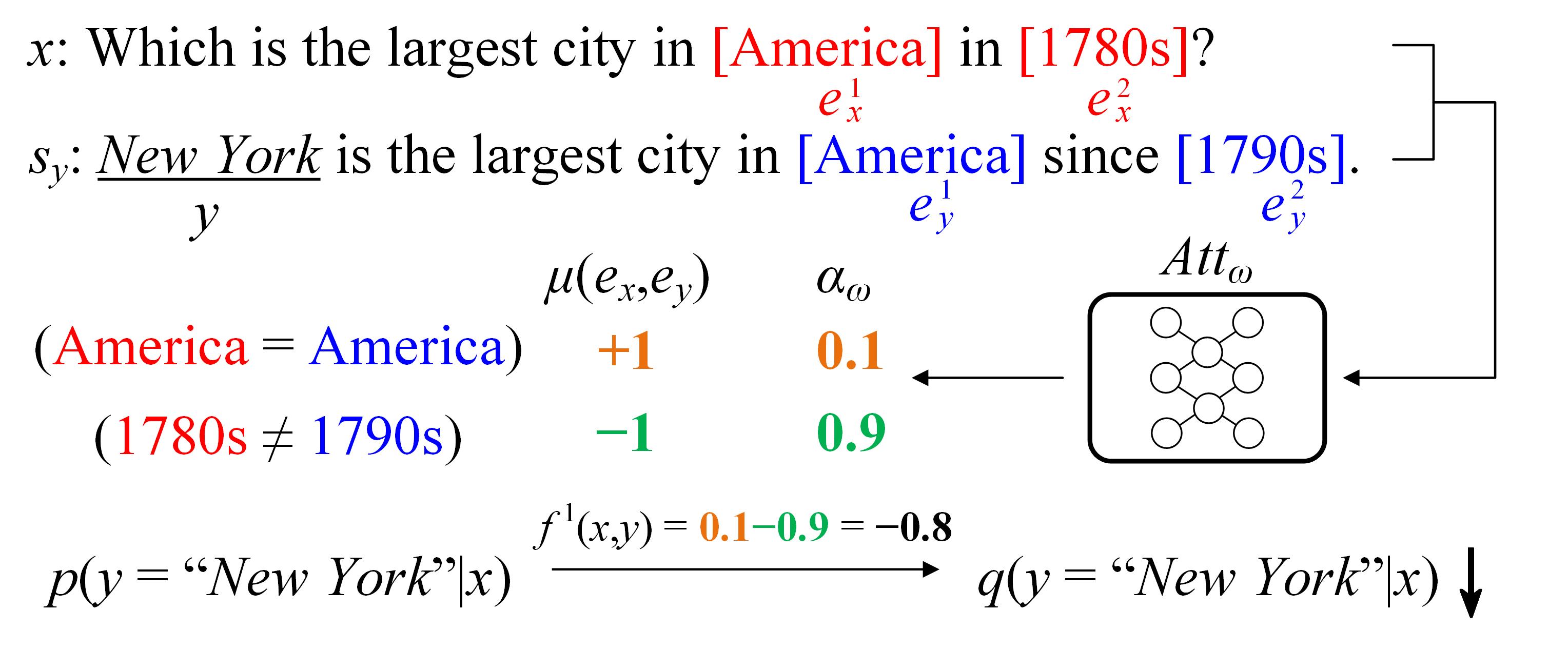}
    \caption{The work flow of entity constraint. The entity pairs are extracted from $x$ and $s_y$, and are weighted by the attention network $Att_{\omega}$. The constraint function $f^1(x,y)$ is then evaluated and used to regularize $q(y|x)$.}
    \label{fig:f}
\end{figure}

\noindent \textbf{Entity Constraint:} The answer should be extracted from a sentence that has the same entities (person, location, time, event, etc) with the question, and on the contrary, the answer should not be extracted from a sentence that has different entities. 
For example, the model should not extract {\it \small``New York"} from sentence {\it \small``New York is the largest city in America since 1790s"} as the answer to the question {\it \small``Which is the largest city in America in 1780s"}.

As shown in Fig.~\ref{fig:f}, we first extract entity pairs\footnote{See section 4.1 {\it Data Preparation}.}  $\{(e_x,e_y)|e_x \in x, e_y\in s_y\}$
between the question sentence $x$ and the sentence $s_y$ where answer $y$ is located
. 
In this example, we have pairs \{(\textit{America}, \textit{America}), (\textit{1780s}, \textit{1790s})\}. 
The entity constraint is formulated as:
$$
 f^{1}(x,y;\omega_1)=\sum_k \alpha_{\omega_1}^k \mu(e^k_x, e^k_y) 
$$
where $\mu(e^k_x,e^k_y)=1$ if the $k^{th}$ entity pair is semantic equivalent or $\mu(e^k_x,e^k_y)=-1$ if semantic different, according to external linguistic knowledge. $\alpha_{\omega_1}=Att(x,s_y;\omega_1)$ is the weight of each entity pair, obtained from an attention network parameterized by $\omega_1$.

Intuitively, $f^{1}(x,y;\omega_1)$ is positive when $y$ is the ground truth, whereas negative when $y$ locates in a semantic different sentence.
Consequently, according to Eq.~\ref{eq:q}, positive $f^1(x,y;\omega_1)$ makes the regularized probability $q(y|x)$ larger than $p(y|x)$. 
Conversely, if $f^1(x,y;\omega_1)$ is negative, $q(y|x)$ becomes smaller. 
\\
\noindent \textbf{Lexical Constraint:} The answer should be extracted from a sentence that has the synonyms (same adjectives/adverbs, full name noun vs. abbreviation, etc.) with the question, or on the contrary, the answer should not be extracted from where antonyms exist.
For example, {\it \small``New York is the largest city in America."} v.s. {\it \small``Which is the smallest city in America?"}.

Similarly, we extract synonym and antonym pairs $\{(w_x,w_y)|w_x \in x, w_y\in s_y\}$ and define the lexical constraint as: 
$$f^{2}(x,y;\omega_2)=\sum_k\alpha_{\omega_2}^k\mu(w^k_x, w^k_y)$$ 
which regularizes the probability of answer $y$ in the same way as $f^{1}(x,y;\omega_1)$. $\alpha_{\omega_2}$ is obtained similar to $\alpha_{\omega_1}$.
\\
\noindent \textbf{Predicate Constraint:} 
Verbs (predicates) sometimes provide crucial semantic information.
For example, in the {\it other neutral case}~\cite{squad2.0} 
irrelevant verbs may act as indicators of irrelevant events. (e.g. {\it \small ``Who discovered Y. pestis?”} vs. {\it \small ``The Black Death was caused by a variant of Y. pestis.”}). 

Since it is difficult to explicitly define whether two verb sequences are semantic equivalent or different,
we model the predicate constraint by a neural network as: $$f^3(x,y;\omega_3)=F(v_x,v_y;\omega_3)$$
where $v_x$ is the verb sequence in the input question $x$ and $v_y$ is the verb sequence in the answer sentence $s_y$ ($y \in s_y$)\footnote{A verb sequence is all the verbs in their original order in the sentence.}. 

The network $F$ is expected to output a positive value for a semantic equivalent pair but a negative value for a semantic different pair. 
Consequently, $f^3(x,y;\omega_3)$ regularizes the answer distribution by decreasing the probability of extracting answers from irrelevant sentences.\\

\subsection{Training Algorithm}
Ganchev et al.~\cite{pr} presented a min-max algorithm to optimize $\mathcal{J}(\theta,q)$ as follows:
\begin{align}
    &E: \quad q^{t+1} = \arg\min_{q} KL(q(y)||p(y|x;\theta^{t})) \\
    &M: \quad\theta^{t+1} = \arg\max_{\theta}E_{q^{t+1}}[\log p(y|x;\theta)]
\end{align}
Hu et al.~\cite{learnedlogicnn} proposed a mutual distillation algorithm that transfers PR into optimization of neural networks. Following the mutual distillation algorithm, we can design the training objectives to update the RC model's parameters $\theta$ and the constraint functions' parameters $\omega=(\omega_1,\omega_2, \omega_3)$.

The RC model $p(y|x;\theta)$ at iteration $t$ is updated with a distillation objective that balances fitting ground truth distribution $g$ (one-hot) and imitating soft predictions of desired regularized distribution $q^{t}$ with distillation parameter $\beta$:
\begin{align}
    \theta^{t+1} = \arg\max_{\theta} &\frac{1}{N}\sum_{i}E_g[\log p(y|x_i;\theta)] \notag \\ 
    &+\beta*E_{q^{t}}[\log p(y|x_i;\theta)]
    \label{eq:theta}
\end{align}

Inspecting the posterior regularization objective, $h(x,y;\omega) = \exp\{C\sum_l\lambda^lf^l(x,y;\omega_l)\}$ should be larger when $y$ is ground truth. In previous work~\cite{learnedlogicnn}, it
was considered as a ``likelihood" metric w.r.t the observations and was optimized the same way as $p(*;\theta)$. In this work, we considered it as a ``score" indicating whether the answer is reasonable or not. 
We labelled positive examples as $\log h^*(x,y)=1$ and negative examples as $\log h^*(x,y)=-1$.
Ground truths or semantic equivalent adversarial sentences are regarded as positive examples, but irrelevant or semantic different adversarial sentences are marked as negative examples.
The parameters of the constraint functions $\omega$ can be optimized using mean-square-error loss (MSE loss):
\begin{align}
    \omega^{t+1} =
    \arg\min_{\omega}&\frac{1}{N}\sum_{i}(
    \log h(x_i,y_i;\omega) - \log h^*(x_i,y_i))^2
    \label{eq:omega}
\end{align}

As $\lambda$ serves to balance different constraints, it is adjusted as follows:
\begin{align}
    \lambda^{t+1} = \arg\max_{\lambda}\frac{1}{N}\sum_{i}E_g[q(y|x_i;\lambda)] 
    \label{eq:lambda}
\end{align}

\begin{table}[ht]
    \normalsize
    \centering
    \begin{tabular}{IlI}
    \hline
    Algorithm: Mutual Distillation \\
    \hline
    Input: data $\{(x_n,y_n)\}^N_{n=1}$ and hyper-parameters $C$,$\beta$\\
    1: Pretrain RC model $p(y|x;\theta)$\\
    2: Initialize constraint functions $f(x,y;\omega)$ and weights $\lambda$ \\
    3: While not converged do:\\
    4: \quad Sample a minibatch (X,Y)\\
    5: \quad (E) Build the desired distribution:\\
    \qquad \qquad $q^{t+1} = p^t\exp\{C\sum\lambda^t f(x,y;\omega^{t})\}$\\
    5: \quad (M) Update ${\theta}$ with distillation objective Eq.~\ref{eq:theta} \\
    6: \quad (M) Update ${\omega}$ with objective Eq.~\ref{eq:omega}\\
    \qquad \qquad and update $\lambda$ with objective Eq.~\ref{eq:lambda}\\
    8: End while\\
    Output: Regularized model $q = p\exp\{C\sum\lambda f\}$\\ 
    \hline
    \end{tabular}
    \label{tab:algorithm}
\end{table}

\section{Experiments}
\subsection{Data Preparation}
We prepared the SQuAD-Adv dataset which consists of original examples, semantic different adversarial (SDA) examples and semantic equivalent adversarial (SEA) examples.
The original examples are randomly sampled from SQuAD1.1 to keep the size balanced with that of adversarial examples. 
SDA examples are randomly sampled from SQuAD-ADDSENT~\cite{squadadv} and unanswerable questions in SQuAD2.0~\cite{squad2.0}. 
We simply generated SEA examples by replacing the adjectives, adverbs, and noun phrases in questions or oracle sentences with their synonyms. 
The statistics of SQuAD-Adv are listed in Table~\ref{tab:dataset}. 
\begin{table}[htbp]
    \normalsize
	\centering
	\caption{\label{tab:dataset}Statistics of SQuAD-Adv which consists of original, SDA, and SEA examples. }
	{
		\begin{tabular}{c I c I c I c I c }
			\hline
			 & \#(Q,P,A) triples & \#Original & \#SDA & \#SEA\\
			\hline
			Train & 120,280 & 45,005 & 51,696 & 23,579 \\
			Test & 23,603 & 9,981 & 8,032 & 5,590 \\
			\hline
		\end{tabular}
	}
	
\end{table}

We used NLTK\footnote{\url{ http://www.nltk.org}} and spaCy\footnote{\url{https://spacy.io}} toolkits to extract entities, verbs, noun phrases, etc.
The words/entities from a question and those from a sentence are paired.
We then filtered irrelevant pairs in which two words/entities do not share similar types or contexts. We obtained
the semantic relationship of each pair, such as {\it``1790s /r/IsA/ nineties"} and {\it ``America /r/Synonym/ U.S."} using  WordNet\footnote{NLTK WordNet interface: {\url{http://www.nltk.org/howto/wordnet.html}}} 
and ConceptNet\footnote{\url{http://conceptnet.io}}.
More details are presented in {\it Supplementary Material}.

\subsection{Experiment Settings}
\label{sec:expset}
We adopted open-source reproduction\footnote{The open-source codes are available at: \url{https://github.com/HKUST-KnowComp/MnemonicReader} and \url{https://github.com/huggingface/pytorch-pretrained-BERT}} of R-Net~\cite{rnet}, Mnemonic Reader (MemReader)~\cite{hu2018reinforced} and BERT~\cite{devlin2018bert:} as the base models. 
The former two models are top performing models on SQuAD except those based on BERT.
BERT introduced a large and empirically powerful language model pre-trained with massive data. It can be fine-tuned to create state-of-the-art models for various tasks including reading comprehension. 

The weight networks $Att(\boldsymbol{x},\boldsymbol{s_y};\omega)$ are used to decide the weights of entity/word pairs in constraint $f^1$ and $f^2$. 
They first apply bi-linear attention to input embeddings of sequence $\boldsymbol{x}=(x_1,...x_m)$ and $\boldsymbol{s_y}=(y_1,..,y_n)$ and then obtain the weight of an entity/word pair by summing the attention scores of some tokens that are included in the pair. 
Concretely, if the two entities of one entity pair $(\boldsymbol{e^k_x},\boldsymbol{e^k_y})$ are represented as $(\boldsymbol{e^k_x}=(x_{s_1},..,x_{e_1}), \boldsymbol{e^k_y}=(y_{s_2},..,y_{e_2}))$ respectively, the weights of the entity pair $\alpha^k$ can be calculated as follows:
\begin{align}
    o^x_i &= Softmax(\sum_{j=1}^{n}y_j^TW_1x_i)\notag \\
    o^y_i &= Softmax(\sum_{j=1}^{m}x_j^TW_1y_i) \notag \\
    \alpha^k &= \sum_{i=s_1}^{e_1}o^x_i + \sum_{j=s_2}^{e_2}o^y_j \notag
\end{align}

While the predicate constraint network $f^3 = F(\boldsymbol{v}_x,\boldsymbol{v}_y;\omega_3)$ adopts one LSTM layer and one feed-forward layer. 
Concretely, $F$ receives two verb sequences and outputs a score as:
\begin{align}
    o^x &= LSTM(\boldsymbol{v}_x;\omega_3) \notag\\
    o^y &= LSTM(\boldsymbol{v}_y;\omega_3) \notag\\
    F(\boldsymbol{v_x},\boldsymbol{v_y};\omega_3) & = tanh(W_3[o^x;o^y;o^x-o^y;o^x*o^y] + b_3) \notag
\end{align}

If the base RC model is BERT, the base model adopts WordPiece embeddings~\cite{wordpiece} and constraint networks share randomly initialized word vectors. 
Otherwise, both base RC models and constraint networks use 300-dimension GloVe word embeddings~\cite{pennington2014glove:} and we kept the pre-trained embeddings fixed during training. 
The dimension of hidden states of $LSTM(\omega_3)$ is set to 100. 
The dimensions of other parameters are $W_1 \in \mathbb{R}^{300\times300}$, $W_3 \in \mathbb{R}^{1\times400}$ and $b_3 \in \mathbb{R}^{1}$. $f^l(\boldsymbol{x},\boldsymbol{y};\omega)$ outputs a scalar for each input.
We used Adam~\cite{Adam} optimizer.
Regularization parameter is set as $C=1$ and distillation parameter is set as $\beta=0.005$.


\begin{table*}[!ht]
    \normalsize
	\centering
	\caption{\label{tab:mainresults}Performance comparison with different settings and on different test subsets. {\it Overall} means the original, SDA and SEA test examples are evaluated together.}
	\setlength{\tabcolsep}{8pt}
	\begin{tabular}{c|c|c| cIc | cIc | cIc | cIc}
		\hline
		\multirow{3}{*}{Base Model}&\multirow{3}{*}{Train set}&\multirow{3}{*}{Method}&\multicolumn{6}{c|}{Test set}&\multicolumn{2}{c}{\multirow{2}{*}{Overall}}\\
		\cline{4-9}
		& & &\multicolumn{2}{c|}{Original} &\multicolumn{2}{c|}{SDA} &\multicolumn{2}{c|}{SEA} &\\
		\cline{4-11}
		& & &EM & F1 &EM&F1 &EM & F1 &EM & F1\\
		\hline
		\multirow{4}{*}{R-Net} & Original
		& Ori-Training   &\bf 65.3 & \bf 74.9& 30.9& 37.9& 54.4 & 65.1 & 51.0 & 60.0 \\
		\cline{2-11}
		&\multirow{3}{*}{SQuAD-Adv} 
		&Adv-Training  & 59.2 & 68.3&57.3&65.6&52.8 & 63.0 & 57.1 & 66.1\\
		\cline{3-11}
		& &Feature-Input  & 60.1 & 69.7 &57.5 & 63.3 & 53.9  & 64.4 & 57.7 & 66.3\\
		\cline{3-11}
        & &PR (ours)  & 61.4 & 69.8  & \bf 59.4 &\bf 70.8& \bf 59.1  &\bf 69.8 &\bf 60.2 &\bf 70.1 \\
		\hline
		\multirow{4}{*}{MemReader}  & Original
		& Ori-Training   &\bf 66.8 &\bf 76.5&35.5&41.9 &56.2 & 67.0  & 53.7 & 62.5\\
		\cline{2-11}
		&\multirow{3}{*}{SQuAD-Adv}
		& Adv-Training  & 59.6 & 69.3 &  58.5& 69.0 & 53.2 & 63.6 &57.7 & 67.8\\
		\cline{3-11}
		& &Feature-Input  & 61.9 & 71.8 & 56.5 & 65.6 & 55.2  & 65.6 & 58.5 & 68.2\\
		\cline{3-11}
        & &PR (ours)  & 62.0 &72.8 &\bf 66.7 & \bf 72.4 & \bf 58.7 & \bf 69.7 & \bf 62.8 & \bf 71.9 \\
		\hline
		\multirow{4}{*}{BERT-base}  & Original
		& Ori-Training   &\bf 73.2 & \bf 82.7 & 41.3 & 48.8 & 62.2  & 73.1 & 59.7 & 68.9\\
		\cline{2-11}
		&\multirow{3}{*}{SQuAD-Adv}
		& Adv-Training  & 71.7 & 79.0 &  76.8 & 84.1 & 68.2 & 76.5 &72.6 & 80.1\\
		\cline{3-11}
		& &Feature-Input  &\multicolumn{8}{c}{Not Applicable}\\
		\cline{3-11}
        & &PR (ours)  & \bf 73.1 & \bf 82.8 &\bf 77.4 & \bf 85.7 & \bf 71.2 & \bf 80.8 & \bf 74.1 & \bf 83.3 \\
		\hline
	\end{tabular}
\end{table*}

We compared the following settings and methods to verify the effectiveness of our PR method:\\
\textbf{Ori-Training}: base RC models are trained with only original examples without any adversarial examples. \\
\textbf{Adv-Training}: base RC models are trained on SQuAD-Adv where both SDA and SEA examples are used in training.\\
\textbf{Feature-Input}: base RC models' inputs are concatenations of word vectors and feature vectors. For each word, its linguistic features such as entity type (``type=DATE") and POS-tag (``pos=NN") are extracted by the toolkit. The feature vector of one word will be formulated as $f(w)\in \mathbb{R}^{|F|}$ where $|F|$ is the amount of features\footnote{In our baselines, $|F|=74$}. Each position $f_i(w)$ use $1/0$ to indicate a property feature like entity type or use a scalar to record a numerical feature like tf-idf. This is a practical method to utilize linguistic knowledge in RC models~\cite{hu2018reinforced}. Feature-Input models are trained on SQuAD-Adv. Due to BERT specifies its input layers, Feature-Input method is only applied to R-Net and MemReader in the following experiments. \\
\textbf{PR}: our method regularizes base RC models with linguistic constraints via PR. PR models are also trained on SQuAD-Adv. Different from {\it Feature-Input}, {\it PR} utilizes linguistic knowledge in the output stage by adjusting the prediction distributions.

To predict unanswerable\footnote{We found that most models learned the strategy to select the ``unanswerable" positions. So we only evaluated with answerable SDA examples in our experiments.} questions, we padded the original document with an extra position (equal to index the answers' positions as $-1$ in BERT) to indicate ``unanswerable". 
We adopted EM (Exact Match) and F1 score as evaluation metrics.

\subsection{Main Results}
For brevity, we presented the following comparisons with respect to the F1 metric of MemReader, but our statements also hold for the EM metric and other two base models.

\textbf{First}, we investigated whether linguistic constraints can benefit RC models. \\
\noindent \textbf{(a-1)} As shown in Table \ref{tab:mainresults}, {\it Ori-Training} achieves good results on the original test examples\footnote{The scores differ from the original papers since we only sampled half of SQuAD1.1 examples for training and the test set is also different, but the codes we used can produce the results reported in the original papers on their own datasets.}, 
but its performance drops remarkably on both SDA (76.5$\rightarrow$41.9) and SEA (76.5$\rightarrow$67.0).
These drops indicate that base RC models cannot handle the over-confidence issue (for SDA examples) or over-sensitivity issue (for SEA examples). \\
\noindent \textbf{(a-2)} Compared to other settings, {\it Ori-Training} performs best on the original test examples but has the worst overall performance because it overfits the original data and lacks of robustness to the adversarial data.\\
\noindent \textbf{(b-1)} {\it Adv-Training} obtains better overall performances compared to {\it Ori-Training} (62.5$\rightarrow$67.8), showing that the model becomes more robust when trained with adversarial examples. \\
\noindent \textbf{(b-2)} Compared to {\it Ori-Training },  {\it Adv-Training} improves the performance on SDA examples (41.9$\rightarrow$69.0) but degrades that on SEA examples (67.0$\rightarrow$63.6). The inconsistency indicates  {\it Adv-Training} cannot balance the over-confidence and over-sensitivity issues simultaneously. \\
\noindent \textbf{(c)} Compared to {\it Ori-/Adv-Training}, our {\it PR} method achieves the best overall performance (62.5/67.8 vs. 71.9), verifying the effectiveness of linguistic constraint regularization. Our method can improve the performance on SDA and SEA examples simultaneously, as the constraints account for two types of adversarial examples at the same time. \\
\noindent \textbf{(d)} Our {\it PR} method improves the robustness of both lightweight models (R-Net and MemReader) and the BERT model, manifesting the linguistic constraint regularization is a versatile method and has a positive effect for all the tested models.

\begin{figure*}[ht]
 \begin{minipage}[t]{0.48\textwidth}
  \centering
     \makeatletter\def\@captype{table}\makeatother
     \caption{\label{tab:sda-train}Performance of Adv-Training (Adv-T) and PR models that were trained with only \textbf{SDA} examples.}
       	\begin{tabular}{c | c | cIc | cIc }
		\hline
		\multirow{3}{*}{Base Model}&\multirow{3}{*}{Method}&\multicolumn{4}{c}{Test}\\
		\cline{3-6}
		&   &\multicolumn{2}{c|}{SDA} &\multicolumn{2}{c}{SEA}\\
		\cline{3-6}
	    &  &EM & F1 &EM & F1  \\
		\hline
		\multirow{2}{*}{R-Net} & Adv-T  &  54.7 & 65.1 &  41.1 &  51.7  \\
& PR &  65.4 &71.8 &\bf 57.2 & \bf 67.5  \\
\cline{2-6}
		\hline
		\multirow{2}{*}{MemReader} & Adv-T  &  58.4 & 69.9 & 41.2 & 52.0 \\
& PR &  65.4 & 74.5 &\bf 57.8 &\bf 68.4 \\
\cline{2-6}
        \hline
        \multirow{2}{*}{BERT-base}  & Adv-T  &  70.1 & 77.8 & 23.8 & 26.4 \\
&  PR &  77.2 & 86.1 &\bf 53.6 &\bf 66.1 \\
\hline
	\end{tabular}
  \end{minipage}
  \quad
  \begin{minipage}[t]{0.48\textwidth}
   \centering
        \makeatletter\def\@captype{table}\makeatother
        \caption{\label{tab:sea-train}Performance of Adv-Training (Adv-T) and PR models that were trained with only \textbf{SEA} examples.}
        \begin{tabular}{c | c | cIc | cIc }
		\hline
		\multirow{3}{*}{Base Model}&\multirow{3}{*}{Method}&\multicolumn{4}{c}{Test}\\
		\cline{3-6}
		&   &\multicolumn{2}{c|}{SDA} &\multicolumn{2}{c}{SEA}\\
		\cline{3-6}
	    &  &EM & F1 &EM & F1  \\
		\hline
		\multirow{2}{*}{R-Net} & Adv-T &  26.7 &  33.4 & 57.9 & 67.9  \\
 & PR & \bf 41.3 & \bf 49.5 & 58.2 & 69.1  \\
		\hline
		\multirow{2}{*}{MemReader}  & Adv-T  &  28.8 &35.5 & 58.1 & 69.1  \\
 & PR & \bf 49.0 & \bf 55.9 & 60.2 & 69.9 \\
        \hline
        \multirow{2}{*}{BERT-base}  & Adv-T  &  42.2 & 48.8 & 70.2 & 79.9  \\
 & PR & \bf 69.8 & \bf 77.2 & 74.3 & 85.9 \\
 \hline
	\end{tabular}
   \end{minipage}
\end{figure*}

\textbf{Second}, we compared the effect of different ways to incorporate linguistic knowledge. As shown in Table~\ref{tab:mainresults}, {\it PR} works more effectively than {\it Feature-Input} when faced with adversarial examples (SDA:65.6$\rightarrow$72.4; SEA:65.6$\rightarrow$69.7). \textbf{For one reason}, regularization on the output distribution is more straightforward than feeding traditional feature vectors into the input layer since the symbolic features may vanish after massive nonlinear operations in neural networks. 
\textbf{For another reason}, {\it PR} applies paired features, which are more informative than features assigned to individual words. For example, feature ``1790s$\neq$1780s" is more useful than features \{``1790s /r/IsA/ DATE", ``1780s /r/IsA/ DATE"\}.

\subsection{Cross Evaluation}
\label{sec:cross}
As discussed above, {\it Adv-Training} models trained with both types of adversarial examples cannot cope with these two types at the same time.
We further conducted experiments to verify whether a model trained with only one type of adversarial (e.g. SDA) examples is robust to the other type of adversarial (e.g. SEA) examples. 

As shown in Table~\ref{tab:sda-train}, if trained with only SDA examples, the performance of {\it Adv-Training} drops remarkably from SDA test examples to SEA test examples (69.9$\rightarrow$52.0). 
Moreover, {\it Adv-Training} performs even worse than {\it Ori-Training} on SEA examples (52.0 v.s. 67.0). 
Similarly, as shown in Table~\ref{tab:sea-train},
if trained with only SEA examples, the performance of {\it Adv-Training} degrades substantially from SEA test examples to SDA test examples (69.1$\rightarrow$35.5), and {\it Adv-Training} is even worse than {\it Ori-Training} when faced with unseen SDA examples (35.5 v.s. 41.9).
The gap is even enlarged when it is applied to the BERT model.
By contrast, {\it PR} performs much better than {\it Adv-Training} on those unseen cases with higher EM/F1 scores and smaller performance gaps between seen and unseen cases. 
Particularly for the BERT model trained on SDA examples where the powerful model may overfit the training data easily, {\it PR} can positively modulate the base RC model substantially (26.4$\rightarrow$66.1).

These results reveal that {\it Adv-Training} mitigates one robustness issue however deteriorates the other issue. Though {\it Adv-Training} improves the robustness of a model, the effect of such a method relies heavily on the training data. 
Considering the larger gaps of BERT's results, where the BERT model leverages very large-scale data with pretraining, it is still insufficient to defend against different adversarial types. 
As SDA examples requires a model to be sensitive enough, while SEA examples requires a model to be confident, it is hard for data-driven models to handle the two cases simultaneously.  
By contrast, our {\it PR} method can handle unseen adversarial types with the help of external linguistic knowledge. The linguistic constraints are applicable to different types of adversarial examples, which consider more symbolic semantics instead of operate in embedding spaces.


\subsection{Ablation Test}

\begin{table}[ht]
    \normalsize
    \centering
    \caption{Ablation test to investigate the effect of different constraints.}
    \begin{tabular}{l|cIc|cIc}
    \hline
         & \multicolumn{2}{c|}{SDA} & \multicolumn{2}{c}{SEA} \\
         \cline{2-5}
         &EM & F1&EM & F1 \\
         \hline
         Adv-Training & 58.5 & 69.0 & 53.2 & 63.6\\
        \hline
    (Full)PR& 66.7 & 72.4 & 58.7 & 69.7 \\
    \hline
    $-$Entity & 62.3 & 71.0 & 57.1 & 67.4\\
    $-$Lexical & 62.5 & 71.6 & 56.2 & 66.5\\
    $-$Predicate & 65.4 & 71.8 & 58.3 & 69.2\\
    \hline
    Only Entity & 61.9 & 70.5 & 56.0 & 66.3 \\
    Only Lexical & 61.2 & 70.2 & 56.9 & 66.8 \\
    Only Predicate& 58.8 & 66.3 & 52.9 & 63.3\\
    \hline
    \end{tabular}
    
    \label{tab:abla_test}
\end{table}

\begin{figure*}[!ht]
    \centering
    \includegraphics[width=18cm]{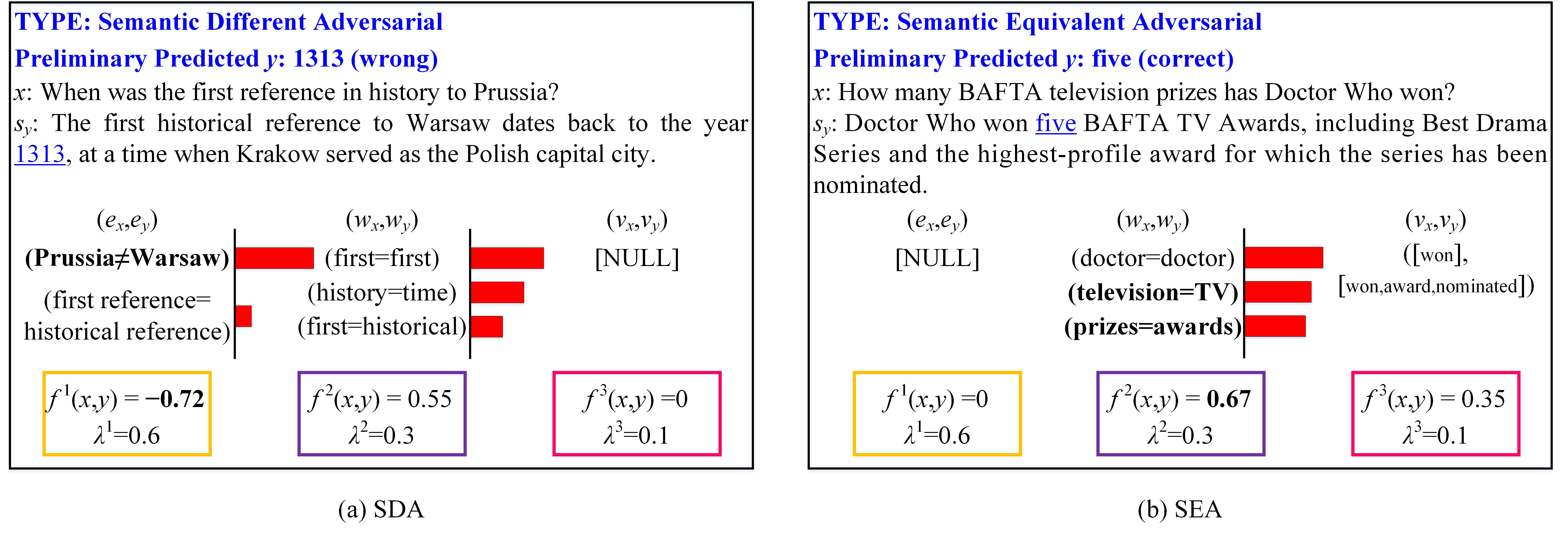}
    \caption{Entity/Word pairs and corresponding weights in SDA and SEA examples. 
    Adversarial pairs (in bold text) were successfully identified by external knowledge and derived expected constraint values $f^l$.
    Note that $\lambda^l$ is only specific to the constraint type thus identical across different question-sentence pairs.}
    \label{fig:case}
\end{figure*}

We conducted ablation test to investigate the effect of each constraint. MemReader was chosen as the base model and it was evaluated on adversarial examples to discriminate between different constraints. 
Results in Table~\ref{tab:abla_test} show that:
\\
\textbf{First}, the entity constraint and lexical constraint play more important roles in robustness than the predicate constraint. On one hand, adversarial examples violating the former two constraints are observed more frequently than those violating the predicate constraint in our data. On the other hand, determining SEA and SDA examples via entity/lexical constraints is more straightforward and precise than predicate constraint.
\\
\textbf{Second}, models with more constraints perform better than those with less constraints: three constraints (Full) $>$ two constraints ($-*$) $>$ single constraint (Only *) $>$ no constraint (Adv-Training). When more constraints are adopted, the performance is better since more adversarial phenomena~\cite{squad2.0} are captured.

\subsection{Case Study and Error Analysis}
We demonstrated here how the constraint weight works in our model via case studies. We also conducted error analysis on failure cases to give insights on the method's behavior.

For the SDA example in Fig.~\ref{fig:case}, our model correctly identifies the most important entity pair (``Prussia$\neq$Warsaw") and obtains a negative $f^1$ value which suppresses the probability of wrong answer {\it ``1313"}.
As for the SEA example, our model successfully identifies the adversarial substitution (``television$=$TV" and ``prizes$=$awards").
The model then derives a positive $f^2$ value to increase the probability of the correct answer. 

In spite of successfully answering these adversarial examples, data processing is still insufficient in our model. 
As shown in Fig.~\ref{fig:case}, pairs like ``history vs. historical" are excluded because we merely compared words with the same POS tags. Verb phrase ``dates back to" and the key entity ``Doctor Who" are not identified by the data processing toolkit either. 

To analyze how the quality of data processing affects the performance, 
we sampled 100 error cases for manual annotation. The reasons why our model failed to answer these adversarial examples broadly fall into the following categories: \\
\textbf{(a)} For $28\%$ examples, the model failed to extract basic verbs, adjectives, etc, and thus had no constraint features. For examples, passive voice ``be colonized" cannot be tagged as a verb. Improving these examples needs more precise tools. \\
\textbf{(b)} For $31\%$ examples, the model failed to group words as an entity like ``15 June 1520", and thus had wrong pairs. Improving these examples needs more rules or better tools in data processing.\\
\textbf{(c)} For $12\%$ examples, the model failed to find the correct relationship of a pair, and thus obtained wrong $f$ values. Improving these examples needs larger knowledge bases.\\
\textbf{(d)} For $29\%$ examples, wrong predictions were just due to the deficiency of models. Better base models and more constraints may benefit these examples.\\

\section{Conclusion}
This paper studies two robustness issues existing in current machine reading comprehension models: over-confidence and over-sensitivity. 
To address these two issues simultaneously, we leverage external linguistic knowledge to impose three linguistic constraints (entity constraint, lexical constraint, and predicate constraint) on the answer distribution via posterior regularization. 
Experiments demonstrate that our method improves the robustness of reading comprehension models, and it is better to cope with these two types of adversarial examples simultaneously.


%

\appendices
\section{Data Processing}
\label{sec:supp-dataprocessing}
We present the details of data processing in this section. 
To obtain the linguistic constraints, we have two steps: the first step is to obtain the entity set (entities), the word set (adjectives, adverbs, noun phrases) and the verb sequence (verbs in original order) for each sentence. The second step is to obtain paired entities and words from the entity or word sets of the two sentences.

The first step has the following procedure:
\textbf{Firstly}, We use NLTK and spaCy toolkits to tokenize a sentence, and then obtain the POS tag and entity type of each word in the sentence.
\textbf{Secondly}, the following rules are applied to decide which set (the entity set, word set, or verb sequence) a word belongs to:
\begin{itemize}

\item Discard a word if its POS tag is in set \{'PDT', 'POS', 'PRP', 'PRP\$', 'RP', 'CD', 'EX'\}.

\item Add a word to the entity set if it has an entity type given by spaCy.

\item Add a word to the verb sequence if its POS tag is VB*.

\item Add a word to the word set if its POS tag is JJ* or RB*.

\item If a word's POS tag is NN*, we first obtain a noun phrases (e.g., 'train station') by merging the word with its adjacent words which have the same POS tag, and then add the phrase to the word set.
\end{itemize}

The second step is to obtain entity or word pairs from the entity and word sets for the input question $x$ and sentence $s_y$. This process is mainly based on the semantic relationship between two words or two entities.
The procedure is as follows:
\textbf{Firstly}, each item (a word or a phrase) in the word set of the question $x$ is paired with each in that of the sentence $s_y$. The same process is applied to the entity sets of $x$ and $s_y$.
\textbf{Secondly}, for each entity or word pair, we decide its semantic relationship sequentially as follows:
\begin{itemize}
    \item If the two items in this pair have different POS tags or entity types, such as (``hot",``city") and (``1949",``America"), the pair is treated as irrelevant and discarded.
    
    \item For each item in the pair, We extract a contextual word set from a sentence with a 10-word window. If the number of overlapping words in the two contextual word sets is less than 3, the pair is treated as irrelevant and discarded.
    
    \item If the two items in this pair are exactly the same, or are defined as synonyms in WordNet, or have relationships like /r/IsA/ and /r/RelatedTo/ according to ConceptNet, the pair is judged as semantic equivalent.
    
    \item For an entity pair, we additionally obtain abbreviations by concatenating initials with ``." like ``United State$\rightarrow$U.S.". If one entity's abbreviation is the same as the other entity, the pair is judged as semantic equivalent.
    
    \item If the two items differ in negative prefix such as ``(unbalanced, balanced), (possible, impossible)", the pair is judged as semantic different.
    
    \item If the two items are defined as antonyms in WordNet, or have /r/Not* relationship according to ConceptNet, the pair is judged as semantic different.
    
    \item For an entity pair, if the two items are different and this pair is not judged as semantic equivalent before, we judge it as semantic different.
\end{itemize}

\section*{Acknowledgment}
This work was supported by the National Key R\&D Program of China (Grant No. 2018YFC0830200),
and partly by the National Science Foundation of China  (Grant No.61876096/61332007).
The authors would like to thank all the workers who help us to annotate sampled data.
Mantong Zhou would like to acknowledge Dr.Yao Xiao and Yijie Zhang for fruitful discussions and data preparation.

\ifCLASSOPTIONcaptionsoff
  \newpage
\fi


\bibliographystyle{plain}
\bibliography{reference}
\end{document}